# Development of a Multiprocessing Interface Genetic Algorithm for Optimising a Multilayer Perceptron for Disease Prediction

Iliyas Ibrahim Iliyas[1], Souley Boukari[2], Abdulsalam Ya'u Gital[3]
[1]Department of Computer Science, University of Maiduguri, Borno State, Nigeria; [2,3]Department of Computer Science, Abubakar Tafawa Balewa University, Bauchi State, Nigeria.
**Correspondence Email:** iliyasibrahimiliyas@unimaid.edu.ng
https://orcid.org/0000-0002-6777-3564

**Abstract**
**Background:** Accurate disease diagnosis enhances effective patient management; however, manual interpretation of complex biomedical data is time-consuming and vulnerable to error. Artificial intelligence systems, particularly machine learning models, can automatically learn complex patterns from high-dimensional clinical and imaging data. The predictive performance of these methods depends critically on proper hyperparameter tuning.
**Methods:** This study introduces a framework that integrates nonlinear feature extraction, classification, and efficient optimisation. First, kernel principal component analysis with a radial basis function kernel reduces dimensionality while preserving 95% of the variance. Second, a multilayer perceptron (MLP) learns to predict disease status. Finally, a modified multiprocessing genetic algorithm (MIGA) optimises MLP hyperparameters in parallel over ten generations. We evaluated this approach on three datasets: the Wisconsin Diagnostic Breast Cancer dataset, the Parkinson's Telemonitoring dataset, and the chronic kidney disease dataset.
**Results:** The MLP tuned by the MIGA achieved the best accuracy of 99.12% for breast cancer, 94.87% for Parkinson's disease, and 100% for chronic kidney disease. These results outperform those of other methods, such as grid search, random search, and Bayesian optimisation. Compared with a standard genetic algorithm, kernel PCA revealed nonlinear relationships that improved classification, and the MIGA's parallel fitness evaluations reduced the tuning time by approximately 60%.
**Conclusion:** The genetic algorithm achieves high computational cost from sequential fitness evaluations, but our multiprocessing interface GA (MIGA) parallelizes this step, slashing the tuning time and steering the MLP toward the best accuracy score of 99.12%, 94.87%, and 100% for breast cancer, Parkison and CKD, respectively.
The built-in graphical user interface then allows clinicians to load data, reduce dimensions, tune hyperparameters, and run predictions without writing code, paving the way for rapid, real-world adoption.
**Keywords:** Multilayer Perceptron, Multiprocessing Interface Genetic Algorithm, Hyperparameter Optimization, Kernel Principal Component Analysis, Parallel Processing

## 1. Introduction

Disease diagnosis is essential for effective patient management and improved clinical outcomes. Clinicians traditionally rely on manual interpretation of laboratory tests and diagnostic imaging to detect diseases. However, this approach is time-consuming and often incapable of capturing the full complexity of high-dimensional biomedical data [1]. Artificial intelligence (AI) is a branch of computer science focused on creating systems capable of performing tasks that require human intelligence. Machine learning (ML) trains algorithms to learn from data and improve their performance without explicit programming [2]. By applying optimization techniques to tune model parameters, AI-driven frameworks automatically extract intricate features from clinical and imaging data, thereby enhancing the predictive accuracy and consistency for diseases [3]. However, the field of AI continues to evolve from rigid rule-based automation to systems that imitate human reasoning, perception, and foresight, allowing machines to take autonomous actions in complex environments. These advancements demonstrate that AI is not merely a tool for efficiency but also a transformative enabler that augments human capabilities across daily life and industry. Hyperparameter tuning is essential in machine learning (ML) because it directly affects model accuracy, generalization, and stability by determining how the learning algorithm behaves during training [4]. Proper tuning helps handle underfitting and overfitting by optimizing parameters or model complexity [5]. Without tuning, models often fail to reach their full predictive potential. Traditional methods such as grid search and random search exhaustively evaluate all possible combinations, making them computationally expensive and inefficient in high-dimensional spaces, a problem known as the curse of dimensionality [6]. These limitations impose more adaptive methods, such as genetic algorithms, that can intelligently search for better hyperparameter combinations. The genetic algorithm (GA) is increasingly used for hyperparameter tuning in machine learning because of its ability to explore large, complex, and nonlinear search spaces more efficiently than traditional methods such as grid or random search. GAs mimic the process of natural selection by encoding hyperparameters as chromosomes, evaluating their performance (fitness), and iteratively improving them through genetic operators such as selection, crossover, and mutation [7]. This evolutionary process enables the GA to avoid local minima and converge toward globally optimal solutions, making it particularly useful for high-dimensional and nonconvex tuning problems. Unlike traditional techniques, which are static and uninformed, the GA is adaptive and uses past evaluations to guide future searches, reducing redundant trials and improving convergence speed. While the GA is effective for exploring large search spaces, it often suffers from slow convergence and high computational cost, especially when complex models are evaluated over many generations [7]. Additionally, its performance is sensitive to the tuning of GA-specific parameters such as population size, crossover, and mutation rates, which can lead to premature convergence or suboptimal results if not well-balanced [7]. This work aims to propose a predictive model for disease prediction via an optimized ML model with an improved genetic algorithm with parallel fitness evaluation called the multiprocessing interface genetic algorithm to increase disease prediction and tuning efficiency.

**Literature Review**

Recent works have shown the effectiveness of the genetic algorithm (GA) in optimizing ML models by addressing the limitations of traditional hyperparameter tuning methods. [8] proposed a hybrid model combining a ResNet-50v2 CNN with a GA for classifying acute lymphoblastic leukemia (ALL) from microscopy images. The GA was used to fine-tune the CNN hyperparameters, resulting in a classification accuracy of 98.46%, outperforming both the random search and Bayesian optimization approaches. [9] developed a GA-aided hyperparameter optimization framework combined with an ensemble learning model to predict respiratory diseases via clinical data. Their methodology also incorporated SHAP-based explainable AI to interpret model predictions. The GA-optimized AdaBoost classifier achieved the highest accuracy among all the models evaluated, showing the efficacy of GAs in tuning complex classifiers and reducing human involvement in parameter selection. In the energy sector, [10] applied a GA to enhance the performance of a random forest model for predicting grid faults. By selecting relevant meteorological features and tuning the model hyperparameters with a GA, the proposed method improved the prediction accuracy by 14.77% over that of standard RF models, illustrating the adaptability of GAs in real-time fault prediction under environmental variability. Furthermore, [11] introduced a multiobjective GA approach for tuning support vector machine (SVM) hyperparameters in imbalanced datasets. Their model, which incorporates decision trees to accelerate GA evaluations, significantly reduces the computational time while improving balanced accuracy metrics such as the G-mean, which are vital for healthcare-related classification tasks. Building on constrained optimization techniques, [12] proposed a novel framework named MLPRGA+5, aimed at configuring multilayer perceptron (MLP) networks through real-coded genetic algorithms. The work implemented advanced genetic operators such as tournament selection with elitism, simulated binary crossover (SBX), and polynomial mutation (PM). The model's efficiency was validated on four UCI datasets, where MLPRGA+5 achieved higher accuracy and reduced network complexity than traditional methods did, confirming the robustness of the evolutionary design in practical neural network tuning. [12] introduced an ensemble model (EGACNN) that stacks a CNN with a GA to optimize hyperparameters such as the dropout rate, batch size, and learning rate. Their method achieved **99.91% accuracy** on MNIST, outperforming the classical CNN and other ensemble models, thereby validating the robustness of GAs in fine-tuning deep neural networks for image classification tasks. [13] optimized an MLP using a GA to predict CKD disease and achieved better accuracy scores of 98.34% and 98.54% for the training and testing processes, respectively.

## 3.0 Methodology

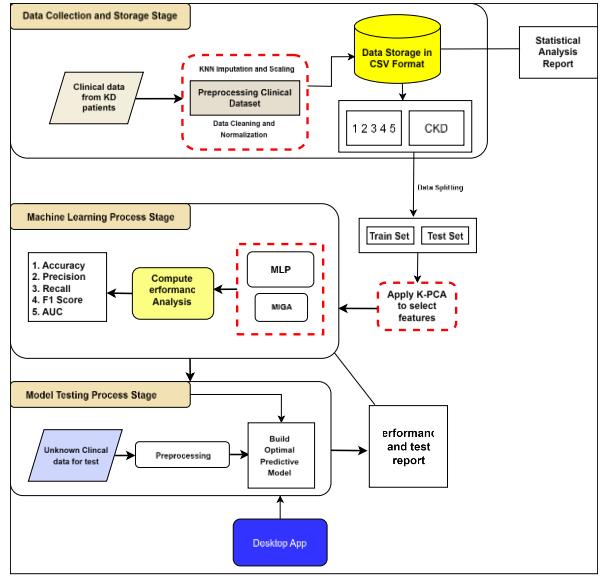

Figure 1: Proposed Framework

### 3.1 Dataset
#### 3.1.1 Chronic kidney disease (CKD)
Chronic kidney disease (CKD) is also called chronic renal failure (CRF). CKD is a clinical syndrome characterized by permanent structural and functional damage to the kidneys that lasts for at least three months and reduces their ability to filter metabolic waste, regulate fluid-electrolyte balance, and control blood pressure (Iliyas et al., 2025). It typically develops gradually and remains unnoticed until approximately 25% of renal function is lost. In adults, CKD is diagnosed when the estimated glomerular filtration rate (eGFR) falls below 60 mL/min/1.73 m² or when the eGFR is ≥ 60 mL/min/1.73 m² in the presence of markers of kidney damage, such as albuminuria over three months or more [14]. Globally, CKD affects more than 800 million individuals and is projected to rank among the five leading causes of death by 2040, highlighting the urgent need for accurate early detection and predictive modelling approaches [15]. The CKD dataset in the study has 400 instances with 25 input features and 1 target feature, and the target feature is called "classification". The dataset has a total of 1009 missing values [16].

#### 3.1.2 Parkison disease
Parkinson's disease (PD) is a progressive neurodegenerative disorder caused by the loss of dopamine-producing neurons in the substantia nigra of the midbrain, leading to hallmark motor symptoms such as resting tremor, bradykinesia, rigidity, and postural instability [17]. Globally, PD affects an estimated 7–10 million people, with the incidence rising markedly after age 60 and a male-to-female prevalence ratio of approximately 1.5:1 [18]. Early in its course, PD often presents with subtle voice- and speech-based changes, and dysphonia is detectable through sustained phonation before overt motor signs appear [19]. Diagnosis remains clinical and is based on established criteria supplemented by neuroimaging and biomarkers, but the insidious onset and overlapping features underscore the need for robust early-detection models trained on specialized PD datasets. The Parkison dataset used in the study is obtained from the UCI ML repository originally collected at Oxford Parkison's Disease Centre. The dataset has a total of 195 records with 23 features and 1 target feature, the target feature is "status", and the dataset also has no missing values [20].

#### 3.1.3 Breast Cancer Disease
Breast cancer is caused by the uncontrolled proliferation of epithelial cells within breast tissue, resulting in the formation of benign or malignant tumours (Islam et al., 2024). Malignant lesions are defined by their capacity to invade adjacent stroma and disseminate to distant organs via lymphatic or hematogenous routes, which critically worsens prognosis. Mammography remains the stage of the art method for early detection; radiographic assessment focuses on lesion morphology, margin characteristics, and tissue density to distinguish benign from malignant findings [21]. Recent diagnostic methods integrate DL techniques to enhance classification performance on large mammographic datasets [22]. The breast cancer dataset used in this study is downloaded from the UCI machine learning repository. The dataset is from the University of Wisconsin Hospital, and it has 569 instances with 32 features, 31 input features and 1 target feature. The target feature is "Diagnosis", and the dataset has no missing values [23].

### 3.2 Data Preprocessing
The study used three datasets: only the CKD dataset had missing values; the missing values were handled via KNN imputation; the categorical features were encoded via one-hot encoding; and the dataset was randomly split into training and testing datasets, with 80% and 20%, respectively.

### 3.3 Kernel-Based Principal Component Analysis
Kernel principal component analysis (kernel PCA) is a type of PCA that extends PCA by mapping data into a high-dimensional feature space via a nonlinear kernel function, such as a radial basis, cosine, sigmoid or polynomial, and then computes principal components in that space, which allows the capture of complex, nonlinear structures that linear PCA cannot detect (Iliyas et al., 2024). This method therefore provides more effective dimensionality reduction for downstream tasks such as disease classification. The study used the radial basis function kernel principal component to identify the best component that explained at least 95% of the total variance.

### 3.4 Multilayer perceptron (MLP)
A multilayer perceptron (MLP) is a type of feed-forward artificial neural network (ANN) comprising an input layer, one or more hidden layers of interconnected neurons, and an output layer; each neuron computes a weighted sum of its inputs plus a bias and applies a nonlinear activation function, enabling the network to approximate complex, nonlinear mappings through gradient-based optimization of its parameters without any feedback loops [24]. By propagating information unidirectionally from input to output, the MLP learns hierarchical feature representations and has demonstrated strong performance in classification tasks across domains and in chronic kidney disease prediction, where MLPs combined with kernel-based dimensionality reduction achieved up to 100% accuracy [3].

### 3.5 Genetic Algorithm
Genetic algorithms (GAs) are a class of population-based metaheuristic optimization techniques inspired by the principles of natural selection and evolutionary biology. A GA begins with a randomly generated population of candidate solutions, known as chromosomes, which are evaluated based on a fitness function that reflects how well each solution solves a given problem. Through iterative processes involving selection, crossover (recombination), and mutation, the algorithm evolves the population over multiple generations to improve overall fitness [25]. The selection phase allows high-performing individuals to pass their genes to the next generation, whereas crossover combines parts of two parents to produce offspring with mixed traits. Mutation introduces diversity by randomly altering genes, helping avoid local optima. GAs are particularly suitable for complex search spaces where traditional gradient-based methods may fail, as they do not require derivative information and are inherently parallelizable [26]. The proposed work modified the GA by enhancing the fitness evaluation with a parallel evaluation through multithreading to increase the computation time. The algorithm of the proposed technique is shown in Algorithm 1.

### 3.5.1 Proposed Multiprocessing Genetic Algorithm (MIGA)
**Algorithm 1** Multi-Processing Interface Genetic Algorithm
**Begin**
1. **Initialize population**
2. Generate an initial population $\mathcal{P}_0 = \{\theta_1, \theta_2, \dots, \theta_P\}$ with random values from $S$
3. **For** each generation, $g = 1$ to $G$:
    a. **Parallel Fitness Evaluation**:
    - Compute $F(\theta_i)$ for each $\theta_i \in \mathcal{P}_g$ by multithreading
    
    b. **Selection**:
    Rank population by fitness
    Select the top 50% as parents $\mathcal{P}_{\text{elite}} \subset \mathcal{P}_g$
    
    c. **Crossover**:
    While $|\mathcal{P}_{g+1}| < P$:
    i. Randomly select two parents $\theta_p^{(1)}, \theta_p^{(2)} \in \mathcal{P}_{\text{elite}}$
    ii. Generate child $\theta_c$ by randomly selecting each gene from either parent
    iii. Append $\theta_c$ to the next generation
    
    d. Mutation
    - For each offspring, with probability $\mu$, one hyperparameter is randomly mutated.
    e. Update population
    - Set $\mathcal{P}_{g+1} \leftarrow$ newly formed generation.
    f. Track the best solution

- If any $\theta \in \mathcal{P}_{g+1}$ has better fitness than $\theta^*$, update $\theta^* \leftarrow \theta$.
3. Return Final best hyperparameter configuration $\theta^*$ and its fitness score $F(\theta^*)$.

End

**4.0 Results**

The study was able to propose a modified GA for optimizing the MLP. The optimized model was used to predict diseases, and the framework was transformed into an app that incorporates the proposed hyperparameter tuning technique to achieve enhanced performance. The study evaluated the performance of the proposed model on three different datasets, namely, breast cancer, CKD and Parkinson's. Table 1 presents the hyperparameters optimized via the genetic algorithm (GA) during the training of the MLPC. The search space for the hidden layer sizes included various configurations ranging from (50), (100), and (150) to two-layer architectures such as (50, 50) and (100, 100). The activation functions considered in the search process were ReLU, tanh, and logistic, allowing the model to explore different nonlinear transformation capabilities. The initial learning rates were tuned across three values: 0.001, 0.01, and 0.1, enabling the GA to adjust the convergence behaviour of the neural network. Finally, the solvers 'adam' and 'sgd' were explored to determine the most effective optimization technique for minimizing loss during training.

Table 1: Hyperparameters Tuned by the Genetic Algorithm

| Hyperparameter | Search Space/Values |
| --- | --- |
| Hidden Layer Sizes | (50), (100), (150), (50, 50), (100, 100) |
| Activation Function | 'relu', 'tanh', 'logistic' |
| Learning Rate Init | 0.001, 0.01, 0.1 |
| Solver | 'adam', 'sgd' |

Table 2 outlines the configuration settings employed for the proposed GA in tuning the hyperparameters of the MLP. The population size was set to 10 individuals, and the algorithm evolved over 10 generations. A mutation rate of 10% was applied, where one hyperparameter in a chromosome was randomly altered. The selection strategy used was elitist selection, retaining the top 50% of individuals based on fitness scores. A uniform crossover approach was adopted to recombine the parent chromosomes. The fitness function was defined as the accuracy score on the test dataset, and evaluations were executed in parallel via ThreadPoolExecutor to accelerate computation. Each chromosome encodes a complete model configuration represented by the tuple [hidden layer size, activation function, learning rate, solver]. The proposed technique was applied specifically to optimize an MLP with a maximum iteration limit of 500.

Table 2: Genetic Algorithm Settings

| GA Parameter | Value |
| --- | --- |
| Population Size | 10 |
| Number of Generations | 10 |
| Mutation Rate | 0.1 (10%) |
| Selection Strategy | Elitist selection (top 50% by fitness) |
| Crossover Strategy | Uniform crossover |
| Mutation Strategy | Random mutation of one hyperparameter |
| Fitness Function | Accuracy score on test set |
| Parallel Evaluation | (ThreadPoolExecutor) |
| Chromosome Representation | [layer size, activation, learning rate, solver] |
| Model Evaluated | MLPClassifier (max_iter = 500) |

Table 3 presents the generationwise performance metrics of the MIGA when applied to the Parkinson dataset. Across 10 generations, the algorithm consistently demonstrated strong optimization capability, with the best accuracy reaching 0.9487 from generation 2 onwards. The minimum and maximum accuracy values for each generation reflect the diversity in model performance within the population, with early fluctuations gradually stabilizing in later generations. The results indicate that the algorithm effectively converges toward high-performing hyperparameter combinations, maintaining the best accuracy of 0.9487 for the majority of the generations.

Table 3: Performance of the MIGA when applied to the MLP on the Parkison dataset

| Generation | Min | Max | Best |
| --- | --- | --- | --- |
| 1 | 0.8718 | 0.9231 | 0.9231 |
| 2 | 0.8974 | 0.9487 | 0.9487 |

| 3 | 0.8974 | 0.9487 | 0.9487 |
| 4 | 0.8974 | 0.9487 | 0.9487 |
| 5 | 0.8718 | 0.9487 | 0.9487 |
| 6 | 0.8974 | 0.9487 | 0.9487 |
| 7 | 0.8462 | 0.9487 | 0.9487 |
| 8 | 0.8462 | 0.9487 | 0.9487 |
| 9 | 0.8974 | 0.9231 | 0.9231 |
| 10 | 0.8974 | 0.9487 | 0.9487 |

Table 4 summarizes the performance of the MIGA across 10 generations on the Breast Cancer dataset. The results show strong classification performance, with the best accuracy, reaching a peak of 0.9912 as early as generation 2 and remaining consistently high throughout subsequent generations. The minimum and maximum accuracies recorded per generation demonstrate moderate variation in early generations, followed by stabilization as the algorithm converges. The sustained best accuracy across multiple generations indicates the robustness and efficiency of the MIGA in evolving optimal hyperparameter configurations for MLP-based classification on the breast cancer dataset.

Table 4: Performance of the MIGA when applied to the MLP on the Breast Cancer dataset

| Generation | Min | Max | Best |
|---|---|---|---|
| 1 | 0.9561 | 0.9625 | 0.9625 |
| 2 | 0.9649 | 0.9912 | 0.9912 |
| 3 | 0.9825 | 0.9912 | 0.9912 |
| 4 | 0.9373 | 0.9825 | 0.9825 |
| 5 | 0.9825 | 0.9825 | 0.9825 |
| 6 | 0.9561 | 0.9825 | 0.9825 |
| 7 | 0.9561 | 0.9825 | 0.9825 |
| 8 | 0.9561 | 0.9825 | 0.9825 |
| 9 | 0.9825 | 0.9825 | 0.9825 |
| 10 | 0.9649 | 0.9912 | 0.9912 |

Table 5 presents the accuracy performance of the MIGA over 10 generations when it is applied to the Chronic Kidney Disease (CKD) dataset. The algorithm consistently achieves exceptional accuracy, with the best performance reaching 1.0000 (100%) from the first generation and maintaining it throughout all generations. The minimum and maximum accuracy values indicate only instability in the population, with minimum values ranging from 0.9625 to 0.9875. This high and stable performance across generations highlights the effectiveness of the MIGA in discovering optimal MLP hyperparameters for CKD classification tasks, suggesting strong model generalizability and convergence.

Table 5: Performance of the MIGA when the MLP is applied to the chronic kidney disease dataset

| Generation | Min | Max | Best |
|---|---|---|---|
| 1 | 0.9625 | 1.0000 | 1.0000 |
| 2 | 0.9750 | 1.0000 | 1.0000 |
| 3 | 0.9750 | 1.0000 | 1.0000 |
| 4 | 0.9750 | 1.0000 | 1.0000 |
| 5 | 0.9750 | 1.0000 | 1.0000 |
| 6 | 0.9750 | 1.0000 | 1.0000 |
| 7 | 0.9750 | 1.0000 | 1.0000 |
| 8 | 0.9875 | 1.0000 | 1.0000 |
| 9 | 0.9875 | 1.0000 | 1.0000 |
| 10 | 0.9750 | 1.0000 | 1.0000 |

Table 6 summarizes the best-performing hyperparameter configurations obtained with the MIGA on the Parkinson's, breast cancer, and chronic kidney disease (CKD) datasets. For the Parkinson dataset, the optimal configuration achieved is 150 hidden layer units, ReLU activation, a learning rate of 0.1, and the Adam solver, which achieves an accuracy of 95.00%. For the Breast Cancer dataset, the proposed model achieved the highest accuracy of 99.12% with 50 hidden units, the tanh activation function, a 0.001 learning rate, and the Adam solver. Additionally, on the CKD dataset, the proposed model performed best with 50 hidden units, tanh activation, a learning rate of 0.1, and the SGD solver, resulting in 100% accuracy. These results demonstrate the adaptability and effectiveness of the MIGA in fitting the MLP configurations for the three different medical datasets.

Table 6: Optimal configuration of hyperparameters applied to each dataset with the MIGA

| Dataset | H/L | Activation | Learning Rate | Solver | Accuracy |
|---|---|---|---|---|---|
| Parkison | 150 | ReLu | 0.1 | Adam | 95.00% |
| Breast Cancer | 100 | Tanh | 0.001 | Adam | 99.00% |
| CKD | 100 | Tanh | 0.1 | Sgd | 100% |

Table 7 shows the timing logs from the hyperparameter tuning experiments. For each of the three datasets, the study recorded the total wall-clock time of the 10-generation GA search both in the standard (single-threaded) mode and with the proposed MIGA (parallel) evaluator and recorded it in seconds.

Table 7: Comparison of the Tuning Time and Speed-Up between the Standard GA and MIGA

| Dataset | Standard GA Time (s) | MIGA Time (s) | Reduction (%) |
|---|---|---|---|
| **Breast Cancer** | 107.05 | 48.05 | 59.0 |
| **Parkison** | 95.30 | 34.20 | 61.1 |
| **CKD** | 71.46 | 11.46 | 60.0 |
| **Average** | 91.27 | 31.24 | 60.3 |

The proposed MLP tuned with the proposed MIGA performed better than the state-of-the-art optimization approaches across all three datasets: on the breast cancer datasets, the summary of the comparison is shown in Table 7.

Table 7: Comparison of MIGA+MLP with several optimization-based models

| Dataset | Author | Model | Optimization Method | Accuracy |
|---|---|---|---|---|
| Breast Cancer | [12] | ANN | RCGAs | 96.00% |
| | [22] | CNN | PSO | 98.23% (DDSM), 97.98% (MIAS) |
| | [27] | LightGBM | PSO | 99.0% |
| | **Proposed Method** | MLP | MIGA | 99.00% |
| CKD | [13] | MLP | GA | 98.54% |
| | [28] | SVM | Grid Search | 99.33% |
| | [9] | MLP | PSO | 92.76% |
| | **Proposed Method** | MLP | MIGA | 100% |
| Parkison | [29]) | SVM | Bayesian Optimization | 92.30% |
| | [17] | MLP | Quantum Particle Swarm Optimization (QPSO) | 93.00% |
| | [19] | NN | GA | 95.00% |

| | Proposed Method | MLP | MIGA | 95.00% |
|---|---|---|---|---|

The study developed a GUI that performs hyperparameter tuning in a unified window: the top panel includes a select file button to load any CSV file; a dropdown to select the target variable; text fields for user hidden-layer sizes, activation functions, learning rates and solvers; and a start-tuning button. The bottom panel displays a scrollable console that logs each generation's best configuration and then summarizes the optimal hyperparameters, MIGA runtime, training time, test accuracy, confusion matrix and classification report. Finally, a save model button exports the trained MLP. Figure 2 shows the interface of the proposed model on the CKD dataset. Figure 3 shows the interface on the Parkinson's dataset. Figure 4 shows the interface on the breast cancer dataset.

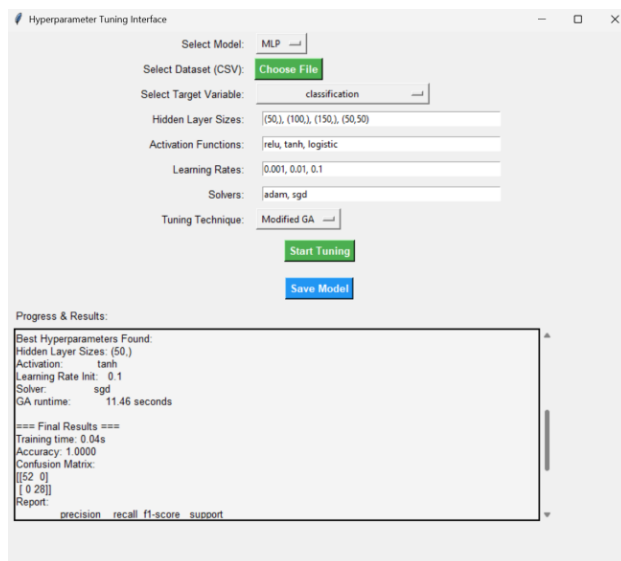

Figure 2: The proposed method interface on the CKD dataset

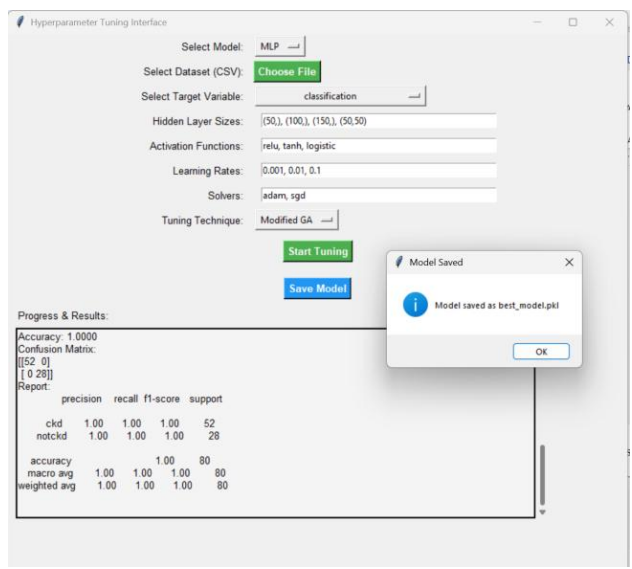

Table 3: Screenshot of the developed interface

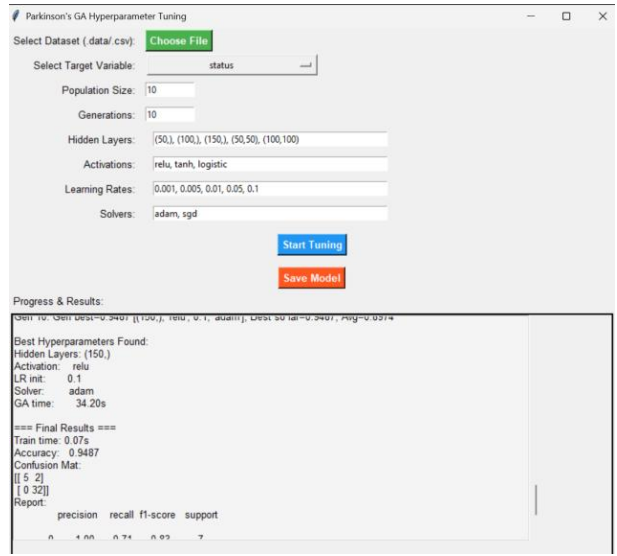

Figure 4: Screenshot of the proposed method interface for Parkinson's disease

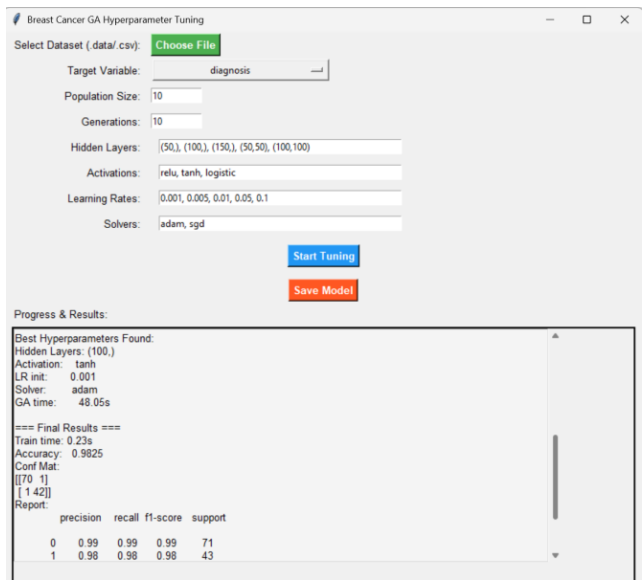

Figure 5: Screenshot of the proposed method interface for treating breast cancer

**5.0 Discussion**
The proposed MIGA overcomes the computational cost of the genetic algorithm by parallelizing fitness evaluations across multiple CPU cores, which reduces the tuning time by approximately 60% compared with that of a single-threaded approach (Ghezelbash et al. 2022). This efficiency gain allows the algorithm to maintain or even expand its population size and number of generations without requiring excessive run times. The combination of nonlinear feature extraction via radial-basis-function kernel PCA and MIGA-tuned multilayer perceptron produced classifiers that generalize exceptionally well, achieving 99.12% for breast cancer, 94.87% for Parkinson's disease and 100% for chronic kidney disease. These results suggest that the framework can capture complex relationships in diverse biomedical datasets. Future work will apply this methodology to additional clinical cohorts to confirm its robustness and to discover dataset-specific feature interactions. The developed graphical user interface, which guides users through data import, dimensionality reduction, hyperparameter optimization and model evaluation without any programming, is crucial for translating these advanced AI methods into everyday clinical practice.

## 6.0 Conclusion

This study introduced a disease-prediction framework that integrates kernel PCA for nonlinear dimensionality reduction, a multilayer perceptron (MLP) classifier, and a modified multiprocessing-enabled genetic algorithm (MIGA) for efficient hyperparameter tuning. The approach was evaluated on three different medical datasets: breast cancer, Parkinson's disease, and CKD. The MLP-tuned MIGA outperforms models optimized with traditional methods, achieving the best accuracies of 99.12% for breast cancer, 94.87% for Parkinson's disease, and 100% for CKD. The use of kernel PCA allows the extraction of nonlinear structures, improving predictive performance, whereas the parallel fitness evaluations in the MIGA reduce the computational time and accelerate convergence toward optimal hyperparameters. A user-friendly GUI further enables nonexpert clinicians to apply the framework to new datasets with minimal effort. This study has proven the potential of combining advanced optimization methods with nonlinear feature extraction to increase disease prediction accuracy and usability in real-world healthcare settings.

**Declarations**

Ethics approval and consent to participate
Not applicable

**Consent for publication**
Not applicable

**Availability of data and materials**
The datasets analysed during the current study are available on the University of California Irvine Repository website and the Kaggle website repository, https://www.kaggle.com/datasets/vikasukani/parkinsons-disease-data-set, https://archive.ics.uci.edu/ml/datasets/Chronic_Kidney_Disease, https://archive.ics.uci.edu/ml/datasets/Breast+Cancer+Wisconsin+(Diagnostic).

**Competing interests**
The authors declare that they have no competing interests.

**Funding**
Not applicable

**Authors' contributions**
II proposed the study conception, design analysis and interpretation; SB provided administrative support and study materials; and AY was a major contributor to manuscript writing. All the authors read and approved the final manuscript.

**Acknowledgements**
Not applicable

**Authors' information**
Department of Computer Science, Faculty of Physical Sciences, University of Maiduguri, Borno State, Nigeria.
Department of Computer Science, Faculty of Computing Science, Abubakar Tafawa Balewa University, Bauchi State, Nigeria